\documentclass[conference]{IEEEtran}
\usepackage[T1]{fontenc}
\usepackage[utf8]{inputenc}
\usepackage{amsmath,amssymb}
\usepackage{booktabs}
\usepackage{graphicx}
\usepackage{multirow}
\usepackage{xcolor}
\usepackage{tikz}
\usetikzlibrary{arrows.meta,positioning,shapes.misc}
\usepackage[hidelinks]{hyperref}
\usepackage{url}
\usepackage{siunitx}
\usepackage{listings}
\graphicspath{{figures/}}

\newcommand{\todo}[1]{{\color{red}[#1]}}
\newcommand{\corpus}{\texttt{eegdash-corpus-v0}}
\newcommand{\generated}[2]{\IfFileExists{figures/#1}{#2}{\todo{figures/#1 not generated yet}}}

\begin{document}
\hypersetup{pdftitle={EEGDash Corpus: A Training-Ready Public Corpus for EEG Foundation Models}}

\title{EEGDash Corpus: A Training-Ready Public Corpus for EEG Foundation Models}

\newif\ifanonymous\anonymousfalse
\newif\ifarxiv\arxivtrue
\ifanonymous
\author{\IEEEauthorblockN{Anonymous authors}\IEEEauthorblockA{Paper under double-blind review}}
\else
\author{\IEEEauthorblockN{Kuber Shahi\IEEEauthorrefmark{1}, Bruno Aristimunha\IEEEauthorrefmark{2}\IEEEauthorrefmark{1} and Arnaud Delorme\IEEEauthorrefmark{1}}
\IEEEauthorblockA{\IEEEauthorrefmark{1}Swartz Center for Computational Neuroscience, University of California San Diego, USA\\
\IEEEauthorrefmark{2}Yneuro, Paris, France\\
\{kshahi, baristimunha, adelorme\}@ucsd.edu}}
\fi

\maketitle
\ifarxiv
{\renewcommand{\thefootnote}{}\footnotetext{\scriptsize Accepted for the Workshop on Machine Learning for EEG Signal Processing (MLESP 2026) at the IEEE International Conference on Bioinformatics and Biomedicine (BIBM 2026). \copyright~2026 IEEE. Personal use of this material is permitted. Permission from IEEE must be obtained for all other uses, in any current or future media, including reprinting/republishing this material for advertising or promotional purposes, creating new collective works, for resale or redistribution to servers or lists, or reuse of any copyrighted component of this work in other works.}}
\fi

\begin{abstract}
Electroencephalography (EEG) foundation models are pretrained on tens of thousands of hours. The largest clinical archives restrict redistribution, while public datasets use different formats and licences. Here, we rebuild the 77 redistributable datasets of the REVE pretraining corpus that reach its 200\,Hz input (14,075 hours across 7,272 subjects) and release them as \corpus{}, with raw-file access by name through EEGDash and a checksummed Zarr store on Hugging Face converted with REVE's preprocessing recipe. Every dataset keeps its licence and any subset of datasets can be selected by name. To test whether this public store pretrains REVE-Base, we train the model twice from scratch with REVE's trainer and score it on seven public OpenEEGBench tasks. First, under linear probing, the frozen encoder reaches 0.331 in chance-adjusted balanced accuracy, against 0.333 for the released model (untrained: 0.200), a difference smaller than the spread between our two pretraining seeds. Second, after fine-tuning, it reaches 0.273 against 0.325 published (0.301 in our rerun), and part of this gap comes from the scoring pipeline. A public, training-ready EEG corpus can therefore pretrain the frozen representations of a foundation model and provide a shared base that other groups can audit.
\end{abstract}

\begin{IEEEkeywords}
EEG, foundation models, self-supervised pretraining, open data, reproducibility, BIDS, EEGDash
\end{IEEEkeywords}

\section{Introduction}

Foundation models for electroencephalography (EEG) learn representations from pooled recordings, and EEG decoding improves with training-data volume~\cite{oliveira2026scaling}.
Published pretraining corpora range from about 2,500 hours~\cite{jiang2024labram} to REVE's 61,415 hours~\cite{elouahidi2025reve}, with recent models reaching about 80,000 hours~\cite{kim2026spera,pan2026labramscale}.
The largest sources require registration and forbid redistribution: four clinical archives hold over 80\,\% of REVE's hours.
Public archives differ in access paths, file formats, channel names, sampling rates and montages.
Fetching, reading, typing, resampling, filtering and segmenting recordings, and checking licences, must precede training.

Choices of filtering, referencing, artifact handling and epoching define a preprocessing multiverse: the same recording yields different training inputs.
Kessler et al.~\cite{kessler2025preprocessing} evaluated 2,592 preprocessing pipelines across seven EEG experiments and found that preprocessing changed EEGNet and time-resolved decoding performance, with effects depending on the experiment and decoder.
Dataset descriptions therefore need to record the processing recipe as well as the source recordings.

Yet groups assemble their corpora by hand from tables of sources.
Models pooling 106 to 521 public and gated datasets release code and weights~\cite{kim2026spera,pan2026labramscale}, but no processed data~\cite{warner2026zuna}; varying the training pool or recipe requires returning to the archives.
REVE's release card names 70 public datasets and provides a processed copy~\cite{reve_dataset_card}.
Auditing that conversion or a dataset's terms still requires the raw archives.

EEG needs a shared, community-maintained data interface, analogous to Hugging Face Datasets~\cite{lhoest2021datasets}, bringing dataset access, licences, versioned descriptions and processing recipes behind a common interface.
Its purpose should be to make alternative pipelines reproducible and comparable, not to prescribe one preprocessing recipe or rebuild access for each experiment.

Here, we rebuild 76 redistributable datasets named in REVE's paper and one more on its release card, all meeting its 200\,Hz input requirement (77 datasets, 14,075 hours), using REVE's recipe.
REVE provides a dataset list, a public trainer and the top OpenEEGBench entry~\cite{guetschel2026openeegbench}.
The result, \corpus{} (\emph{the corpus} below), combines raw-file access and verification by name through EEGDash~\cite{aristimunha2026eegdash} with a checksummed Hugging Face store whose dataset-scoped buckets keep their licences and can be selected independently.

Our three contributions (Fig.~\ref{fig:concept}) are to audit the 77 redistributable datasets, retaining an estimated 79\,\% of REVE's tokens; to package them into 632 Zarr buckets (3.6\,TB) that stream into the trainer; and to retrain REVE-Base twice from scratch.
Its frozen encoder matches the released model within the spread of the two seeds across seven public OpenEEGBench tasks, although the released model was trained on 4.4 times as many hours.
Rerunning the released weights also measures how much of the fine-tuning gap comes from the scoring pipeline.

\begin{figure}[t]
\centering
\definecolor{pref}{HTML}{006CA3}\definecolor{pours}{HTML}{F7941D}\definecolor{pfloor}{HTML}{7F7F7F}
\begin{tikzpicture}[
  x=1cm, y=1cm, font=\fontsize{6.5}{7.5}\selectfont, line cap=round,
  hd/.style={anchor=base west, inner sep=0pt, font=\scriptsize\bfseries, text=black!85},
  cell/.style={draw=#1, line width=0.55pt, fill=#1!6, rounded corners=1.2pt, anchor=west, align=left,
    inner xsep=2pt, inner ysep=1.8pt, minimum height=0.5cm, text width=1.66cm, text=black!90,
    font=\fontsize{6.5}{7.2}\selectfont},
  cell/.default=black!45,
  plaincell/.style={draw=none, fill=none, anchor=west, align=left,
    inner xsep=2pt, inner ysep=1.8pt, minimum height=0.5cm, text width=1.66cm, text=black!90,
    font=\fontsize{6.5}{7.2}\selectfont},
  same/.style={anchor=west, inner sep=0pt, xshift=3pt, font=\fontsize{6.5}{7.5}\selectfont\itshape, text=black!50},
  rowlab/.style={anchor=west, inner sep=0pt, align=left, text=black!90},
  tst/.style={anchor=west, inner sep=0pt, align=left, text width=1.45cm, text=black!85}]
\def\ca{1.42}\def\cb{3.32}\def\cc{5.22}\def\cd{7.15}
\node[hd] at (0,0.16) {model};
\node[hd] at (\ca,0.16) {pretrained on};
\node[hd] at (\cb,0.16) {weights};
\node[hd] at (\cc,0.16) {scored with};
\node[hd] at (\cd,0.16) {tests};
\draw[black!30, line width=0.4pt] (0,0.06) -- (8.6,0.06);
\foreach \y in {-0.62,-1.27,-1.92} \draw[black!12, line width=0.3pt] (0,\y) -- (8.6,\y);
\fill[pref] (0.1,-0.3) circle (0.075);
\node[rowlab] at (0.28,-0.3) {published};
\node[plaincell] at (\ca,-0.3) {92 datasets,\\61,415\,h};
\node[plaincell] at (\cb,-0.3) {released\\REVE-Base};
\node[plaincell] at (\cc,-0.3) {leaderboard};
\node[tst, text=black!55] at (\cd,-0.3) {the reference};
\draw[pref, line width=0.7pt, fill=white] (0.1,-0.95) circle (0.07);
\node[rowlab] at (0.28,-0.95) {our rerun};
\node[same] at (\ca,-0.95) {same};
\node[same] at (\cb,-0.95) {same};
\node[cell=pref] at (\cc,-0.95) {benchmark\\package};
\node[tst] at (\cd,-0.95) {the scoring\\pipeline};
\fill[pours] (0.1,-1.6) circle (0.075);
\node[rowlab] at (0.28,-1.6) {retrained};
\node[cell=pours] at (\ca,-1.6) {the corpus,\\14,075\,h};
\node[cell=pours] at (\cb,-1.6) {REVE's code,\\two seeds};
\node[same] at (\cc,-1.6) {same};
\node[tst] at (\cd,-1.6) {the corpus};
\fill[pfloor] (0.1,-2.25) circle (0.075);
\node[rowlab] at (0.28,-2.25) {untrained};
\node[cell=pfloor] at (\ca,-2.25) {none};
\node[cell=pfloor] at (\cb,-2.25) {random init.};
\node[same] at (\cc,-2.25) {same};
\node[tst, text=black!55] at (\cd,-2.25) {the floor};
\draw[black!30, line width=0.4pt] (0,-2.62) -- (8.6,-2.62);
\node[anchor=north west, inner sep=0pt, text=black!85] at (0,-2.72) {all rows: the same seven public OpenEEGBench tasks, none of them in the corpus};
\end{tikzpicture}
\caption{\textbf{Rerunning the released weights checks the scoring pipeline; retraining REVE-Base on the corpus tests the corpus.} One row per scored model: the released REVE-Base as published (filled blue) and as rerun with the benchmark package (hollow blue), REVE-Base retrained on the corpus (orange), and the untrained encoder (grey). A box marks what a row changes against the row above it; \emph{same} is held fixed.}
\label{fig:concept}
\end{figure}

\section{Background}

\textbf{EEG foundation models and their data.} BIOT~\cite{yang2023biot} cuts each channel into fixed-length segments, so that different channel sets can share one transformer. LaBraM~\cite{jiang2024labram} and CBraMod~\cite{wang2025cbramod} pretrain on about 2,500 and more than 9,000 hours, respectively, and EEGPT~\cite{wang2024eegpt} on five datasets. REVE~\cite{elouahidi2025reve} takes electrode coordinates as input; its released position bank supplies coordinates for 543 known electrode names. It accepts different montages when positions are available. Dense batches share a channel count and window length, but channel identities may differ. REVE was pretrained by masked autoencoding of 10-second samples from 92 datasets, most of its hours clinical recordings from the Temple University Hospital (TUH) corpus~\cite{obeid2016tuh} and I-CARE. This lets the corpus retain different native EEG channel sets with resolvable positions rather than reducing all datasets to one common montage (Section~\ref{sec:recipe}).

\textbf{Archives and access.} The Brain Imaging Data Structure (BIDS)~\cite{gorgolewski2016bids} and its EEG extension~\cite{pernet2019eegbids} fix how a dataset lays out its directories and sidecar metadata, but allow several signal formats. OpenNeuro~\cite{markiewicz2021openneuro} and the NeuroElectroMagnetic data Archive and tools Resource (NEMAR)~\cite{delorme2022nemar} archive BIDS datasets. EEGDash~\cite{aristimunha2026eegdash} catalogues the datasets of both archives and turns any of them into a braindecode~\cite{schirrmeister2017braindecode} dataset. A query returns one record per recording, with its channel names, sampling rate, and duration, before anything is downloaded. MOABB~\cite{jayaram2018moabb} does the same for brain--computer-interface (BCI) datasets, and sixteen of these reach the corpus through NEMAR. The corpus builds on this catalogue: EEGDash resolves each dataset to its files, and the corpus adds the conversion and the store.

\textbf{Benchmark.} OpenEEGBench~\cite{guetschel2026openeegbench} evaluates frozen or fine-tuned encoders on twelve downstream tasks, with fixed splits and one training schedule, and keeps a public leaderboard.

\section{From REVE's Corpus to a Public Store}
\label{sec:corpus}

\subsection{Seventy-Seven Redistributable Datasets Capture \texorpdfstring{79\,\%}{79\%} of REVE's Tokens}

The store holds the openly redistributable part of REVE's pretraining corpus, and every dataset in it traces back to a public deposit. To establish that provenance, we resolved each of the 89 datasets named in the REVE paper (92 datasets and 61,415 hours in all~\cite{elouahidi2025reve}) to a deposit in the EEGDash catalogue and the source archives, and recorded for each its identifier, licence and source (Appendix~\ref{app:excluded}). Four are clinical archives released under registration or a data use agreement: the TUH corpus~\cite{obeid2016tuh}, the Harmful Brain Activity Classification set, the Two Decades Brainclinics Research Archive (TDBRAIN), and I-CARE~\cite{bdsp_icare,amorim2023icare}. Together, they hold about 50,000 hours (over 80\,\% of REVE's total). Eight further BCI datasets were left out under this release's licence policy: six carry no-derivatives terms, and two had no explicit redistribution licence at the sources checked. One more, Zhou2016~\cite{zhou2016mi}, is left out because the deposit REVE's card names (\texttt{nm000226}) is a 100\,Hz copy, below the recipe's 200\,Hz floor; NEMAR's other deposit of the same data (\texttt{nm000115}, the one MOABB uses) holds the original 250\,Hz recordings, from which a later version of the store can take it. The release contains 76 datasets named in the paper, and REVE's release card~\cite{reve_dataset_card} adds one dataset that the paper does not name. Together, these 77 datasets (56 from OpenNeuro and 21 from NEMAR) make up the store.

\begin{figure*}[t]
\centering
\generated{corpus_bubbles.pdf}{\includegraphics[width=\textwidth]{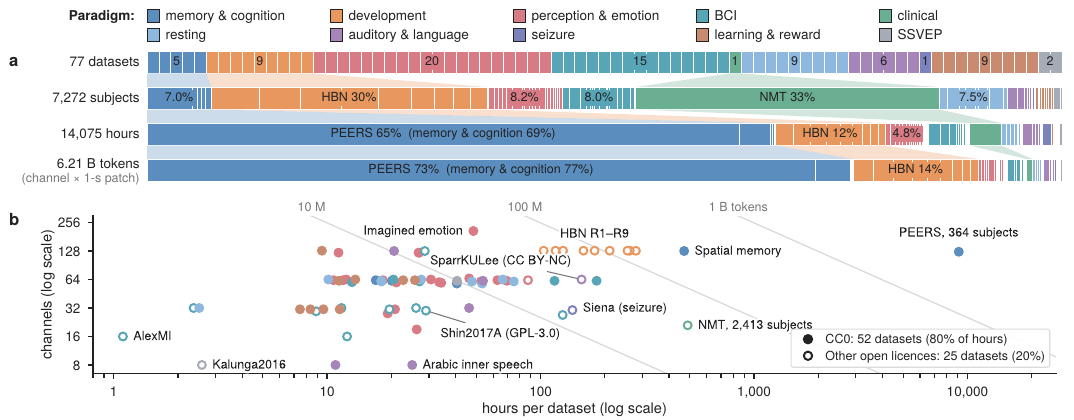}}
\caption{\textbf{One dataset, PEERS, holds 65\,\% of the hours and 73\,\% of the tokens; NMT and the nine Healthy Brain Network releases hold 63\,\% of the subjects.} (a)~Share of the datasets, subjects, hours and tokens per paradigm (colour); each block is one dataset, and ribbons follow PEERS, the Healthy Brain Network and NMT across rows. (b)~Hours and duration-weighted mean channels per dataset (log axes); grey lines mark equal token counts. Filled: CC0 (52 datasets, 80\,\% of the hours); hollow: the other open licences. Per-dataset numbers in Table~\ref{tab:datasets}.}
\label{fig:bubbles}
\end{figure*}

The resulting distribution across sources and paradigms is uneven (Fig.~\ref{fig:bubbles}). A single free-recall study, PEERS~\cite{peers_ds004395}, holds 65\,\% of the hours, while the NMT scalp EEG archive~\cite{khan2022nmt} and the nine Healthy Brain Network releases~\cite{alexander2017hbn} supply 63\,\% of the subjects. The other 66 datasets span a variety of research studies and BCI recordings, with one seizure set~\cite{siena2020} among them (Appendix~\ref{app:datasets}). Memory and cognition (PEERS and four smaller studies) accounts for 69\,\% of the hours; the corpus web page\footnote{\url{https://eegdash.org/corpus/}} gives the full breakdown by paradigm.

Because a channel-wise model consumes patches per electrode rather than raw duration, hours alone undercount what the trainer receives. REVE cuts every sample into one token per channel and per 1-second patch, with a new patch every 0.9 seconds, so a sample of $C$ channels holds $11C$ tokens and a dataset's tokens grow with its hours times its channels (Fig.~\ref{fig:bubbles}b). Counted this way, the corpus holds 6.21 billion tokens per epoch across 112 duration-weighted mean channels: 4.53 billion (73.0\,\%) from PEERS at 126 mean channels, 0.85 billion (13.7\,\%) from the nine Healthy Brain Network releases at 129 mean channels, 0.57 billion (9.2\,\%) from the other 46 OpenNeuro datasets at 83 mean channels, and 0.26 billion (4.1\,\%) from the 21 NEMAR datasets at 42 mean channels. A token from a 129-channel research cap and one from 19-channel clinical monitoring need not be worth the same downstream. Applied to the channel-range table of REVE's paper, the same formula gives an estimated 7.9 billion tokens for its 61,415 hours. The corpus thus holds 23\,\% of the paper's hours and roughly 79\,\% of its tokens (Appendix~\ref{app:compare}).

\subsection{One Conversion Recipe Replicates REVE's Training Inputs}
\label{sec:recipe}

To verify that the store holds what REVE's trainer would have read, we pass every dataset through the same path and apply REVE's conversion recipe (Fig.~\ref{fig:pipeline}). We fetched each dataset through EEGDash, whose cache follows the archive's own storage paths; for OpenNeuro a direct mirror of the storage bucket gave the same tree, allowing the converter to reopen every file offline.

\begin{figure}[t]
\centering
\input{figures/pipeline_traces.tex}%
\definecolor{pblue}{HTML}{006CA3}%
\definecolor{porange}{HTML}{F7941D}%
\definecolor{pbox}{HTML}{F7F9FC}%
\begin{tikzpicture}[
  x=1cm, y=1cm, font=\scriptsize, line cap=round, line join=round,
  lab/.style={align=left, anchor=north west, inner sep=0pt, text=black!85, text width=#1},
  lab/.default=3.60cm,
  small/.style={font=\fontsize{6}{7}\selectfont},
  cardtxt/.style={font=\fontsize{5.8}{6.8}\selectfont, align=center, inner sep=0pt, text=black!90},
  note/.style={font=\fontsize{5.8}{6.8}\selectfont, text=black!65},
  actor/.style={align=left, anchor=north west, inner sep=0pt, font=\scriptsize\bfseries, text=black!90},
  role/.style={align=left, anchor=north west, inner sep=0pt, font=\fontsize{6}{7}\selectfont\itshape, text=black!65},
  src/.style={draw=pblue!65, fill=pblue!6, rounded corners=1.2pt, inner xsep=2.5pt, inner ysep=1.6pt, align=center, font=\fontsize{6}{7}\selectfont, text=black!85},
  bcard/.style={draw=black!35, fill=white, rounded corners=1.2pt, line width=0.45pt},
  flow/.style={-{Latex[length=1.3mm,width=1.1mm]}, black!55, line width=0.5pt},
  trace/.style={line width=0.22pt},
  sig/.style={shift={(0,#1)}, x=\dw*0.01mm, y=0.0055mm}]
\def\dx{1.72}\def\dw{2.78}\def\tx{4.74}
\node[src] (on) at (\dx+0.22,0.02) {\textbf{OpenNeuro} (56 datasets)};
\node[src] (nm) at (\dx+2.70,0.02) {\textbf{NEMAR} (21 datasets)};
\draw[flow] (on.south) -- (\dx+0.22,-0.38);
\draw[flow] (nm.south) -- (\dx+2.70,-0.38);
\node[actor, text=pblue] at (0,-0.40) {1.\ EEGDash};
\node[role] at (0,-0.66) {fetch \& verify};
\begin{scope}[shift={(\dx,-0.52)}]
  \foreach \t/\y in {A/0,B/-0.13,C/-0.26,D/-0.39,E/-0.52,G/-0.78}
    {\draw[trace, black!65, sig=\y] plot coordinates {\csname reveRaw\t\endcsname};}
  \foreach \t/\y/\n in {F/-0.65/CB1,H/-0.91/Trigger}
    {\draw[trace, black!42, sig=\y] plot coordinates {\csname reveRaw\t\endcsname};
     \node[note, anchor=east, inner sep=0pt] at (-0.22,\y) {\n};}
  \foreach \y in {0,-0.13,...,-0.92}
    {\draw[black!35, densely dotted, line width=0.4pt] (-0.18,\y) -- (-0.04,\y) (\dw+0.04,\y) -- (\dw+0.18,\y);}
\end{scope}
\node[lab] at (\tx,-0.38) {\textbf{Recordings as deposited}\\
  \textcolor{black!65}{\fontsize{5.8}{6.6}\selectfont Example: 65 ch., 1,000\,Hz. Grey: CB1 (no coordinates), Trigger (non-EEG).}\\[1pt]
  {\fontsize{5.8}{6.6}\selectfont 77 datasets fetched, each file checked by SHA\mbox{-}256}};
\begin{scope}[yshift=-0.20cm]
\draw[black!20, line width=0.4pt] (0,-1.66) -- (8.35,-1.66);
\node[actor] at (0,-1.80) {2.\ Converter};
\node[role] at (0,-2.06) {REVE's recipe};
\draw[flow] (\dx+1.39,-1.32) -- (\dx+1.39,-1.86);
\begin{scope}[shift={(\dx,-2.0)}]
  \foreach \t/\y in {A/0,B/-0.13,C/-0.26,D/-0.39,E/-0.52,F/-0.65}
    {\draw[trace, black!85, sig=\y] plot coordinates {\csname reveTrace\t\endcsname};}
  \draw[black!80, line width=0.45pt] (0,0.1) -- (0,-0.75) (\dw,0.1) -- (\dw,-0.75);
  \draw[black!75, line width=0.4pt] (0,-0.86) -- (\dw,-0.86) (0,-0.80) -- (0,-0.92) (\dw,-0.80) -- (\dw,-0.92);
  \node[small, anchor=north, inner sep=0pt] at (\dw/2,-0.94) {one 10\,s sample $=$ 2,000 time points};
\end{scope}
\node[lab] at (\tx,-1.80) {\textbf{Standardised 10\,s samples}};
\node[lab, font=\fontsize{6}{7.2}\selectfont, text width=3.75cm] at (\tx,-2.04) {%
  \textbullet\ Keep EEG ch.\ in REVE's position bank\\
  \textbullet\ Band-pass 0.5--99.5\,Hz\\
  \textbullet\ Resample to 200\,Hz; cut 10\,s samples\\
  \textbullet\ Refuse OpenEEGBench source datasets\\[1pt]
  \textcolor{black!65}{\fontsize{5.8}{6.6}\selectfont 62 ch.\ kept (example); 418 parts audited}\vspace{-1pt}};
\draw[black!20, line width=0.4pt] (0,-3.52) -- (8.35,-3.52);
\node[actor] at (0,-3.66) {3.\ Corpus};
\node[role] at (0,-3.92) {store};
\node[note, align=left, anchor=north west, inner sep=0pt, text=black!75] at (0,-4.22) {632 buckets:\\one per dataset\\$\times$ electrode set};
\draw[draw=porange!85, fill=porange!6, line width=0.55pt, rounded corners=1.8pt] (1.42,-5.72) rectangle (6.55,-3.66);
\node[anchor=north west, inner sep=0pt, font=\fontsize{5.8}{6.8}\selectfont, text=black!85] at (1.52,-3.74) {\textbf{\textcolor{porange!60!black}{One Zarr store on Hugging Face}} \textcolor{black!65}{(3.6\,TB; 632 buckets)}};
\draw[flow] (\dx+1.39,-3.20) -- (\dx+1.39,-3.66);
\draw[bcard] (1.52,-5.12) rectangle (3.12,-4.14);
\node[cardtxt] at (2.32,-4.63) {\textbf{HBN R9}\\(10\,445, 2000, 129)\\\textcolor{black!60}{CC-BY-SA}};
\draw[bcard] (3.19,-5.12) rectangle (4.79,-4.14);
\node[cardtxt] at (3.99,-4.63) {\textbf{Perceive/imagine}\\(24\,987, 2000, 62)\\\textcolor{black!60}{CC0}};
\draw[bcard] (4.86,-5.12) rectangle (6.46,-4.14);
\node[cardtxt] at (5.66,-4.63) {\textbf{Kalunga2016}\\(920, 2000, 8)\\\textcolor{black!60}{CC-BY}};
\node[anchor=north west, inner sep=0pt, align=left, text width=5.0cm, font=\fontsize{5.8}{6.8}\selectfont, text=black!75] at (1.52,-5.22)
  {$(N, 2000, C)$ = samples, time points, channels;\\
   with session mean/s.d., coordinates, licence};
\node[draw=black!45, fill=pbox, rounded corners=1.2pt, inner xsep=2.2pt, inner ysep=2.2pt, align=center, anchor=west, font=\fontsize{5.8}{6.8}\selectfont, text=black!85] (train) at (6.88,-4.63) {\textbf{REVE trainer}\\reads any subset\\of buckets};
\draw[flow] (6.55,-4.63) -- (train.west);
\end{scope}
\end{tikzpicture}
\caption{\textbf{Each recording is fetched as deposited, converted once with REVE's recipe and stored in a bucket that keeps its dataset's channel set and licence.} One example recording (Perceive/imagine, participant~1, 70--80\,s) is followed through the pipeline. Top: EEGDash fetches the 77 datasets from OpenNeuro and NEMAR as deposited and checks every file against its SHA-256. Here, 65 channels at 1,000\,Hz, offsets removed for display; grey traces are a channel without a position in REVE's bank (CB1) and a non-EEG channel (Trigger). Middle: the converter keeps the EEG channels that resolve in REVE's position bank, band-passes, resamples to 200\,Hz, cuts 10-s samples of 2,000 time points and refuses the benchmark's source datasets. Bottom: 632 buckets, one per dataset and electrode-name set (three shown); each is an $(N, 2000, C)$ array with session mean and s.d., electrode coordinates and licence. REVE's trainer reads any subset of buckets.}
\label{fig:pipeline}
\end{figure}

Every recording is then processed with REVE's recipe~\cite{elouahidi2025reve}. We retain channels typed as EEG and resolve their names against REVE's electrode-position bank, while dropping channels without a known position and skipping recordings of 10 seconds or less or sampled below 200\,Hz. We band-pass filter the signal at 0.5--99.5\,Hz, resample it to 200\,Hz, and store it as float32 microvolts alongside each session's per-channel mean and standard deviation for z-scoring at load time. Native montages and references are preserved, so a 170-channel PEERS session and an 8-channel motor-imagery (MI) session share one store. We adopted REVE's recipe so that the corpus can pretrain that model, and added nothing to it (no notch filter, no re-referencing, and no artefact rejection).

We ran the conversion as 418 parts of at most 200 hours or 400 recordings each, and merged the audited parts into one store. To confirm that the resulting arrays are identical to those of REVE's memory-mapped layout, we converted two datasets (26,342 samples over 35 montages) both ways from the raw files. Every sample, statistic, and position came out identical, and REVE's training script produced the same number of batches from the same samples. The source datasets of the benchmark's tasks are refused by identifier at planning, at conversion, and at the merge, so no corpus dataset shares a DOI, a source address, or a title with them.

\subsection{Dataset-Scoped Buckets Enable Direct Subset Training}
\label{sec:store}

To isolate individual datasets so that a trainer can select any subset without rewriting the store, we organise the converted recordings into 632 Zarr~\cite{zarr} buckets. Each bucket holds the recordings of one dataset that share one exact set of electrode names (one bucket for a single-montage dataset, several for a multi-montage dataset, and never a bucket across two datasets). Each bucket's \texttt{data} array has shape $(N, 2000, C)$, one row per 10-second sample at 200\,Hz, stored next to the session index of each sample, the per-session mean and standard deviation, the electrode positions in metres, and the dataset's identifier, licence, and source. One sample is one chunk and 128 samples form one compressed shard file, so a training step reads only the samples it needs, and lossless compression brings the store to 3.6\,TB (0.8 of its float32 size). Because no bucket spans two datasets, any subset of datasets is a list of buckets, and the store grows by adding buckets. This makes two operations that are costly when datasets are scattered across archives and formats~\cite{shi2026registry} cheap here: scaling a pretraining corpus by adding datasets, and aligning datasets, that is, bringing recordings from different studies into one layout so that they can be pooled in a single training run. Pooling more datasets in pretraining has been shown to raise downstream scores~\cite{zhang2025mdjpt,chen2026eegarena}, and in this store it is a change to the bucket list. REVE's patched loader takes the identifiers of the datasets to leave out as an argument (Section~\ref{sec:validation}), and because the stored values are band-passed microvolts, models that normalise differently can do so at load time.

\subsection{Per-Dataset Licences Preserve Source Terms Under One Name}

We give the corpus a single identifier so that any pipeline can fetch and verify every file by name (Fig.~\ref{lst:open}). EEGDash resolves \corpus{} to the raw files of the 77 datasets, downloads them, and checks each file against its hash. The converted store is the Hugging Face dataset \texttt{EEGDash/eegdash-corpus-v0}\footnote{\url{https://huggingface.co/datasets/EEGDash/eegdash-corpus-v0}; code: \url{https://github.com/eegdash/eegdash-corpus}.}, which ships with the licence table and bibliography of the 77 datasets, the bucket table, the per-session manifest, and the SHA-256 of every file (with build and scoring code under BSD-3-Clause).

\begin{figure}[t]
\begin{lstlisting}
from eegdash import EEGDashCorpus

corpus = EEGDashCorpus("eegdash-corpus-v0")
corpus.download()
for raw in corpus.records():
    ...
\end{lstlisting}
\caption{\textbf{One name fetches the raw files of all 77 datasets.} \texttt{records()} yields one recording at a time.}
\label{lst:open}
\end{figure}

Rather than imposing a single wrapper licence, the store preserves the licence of each source deposit in the attributes of its buckets, and the licence table that ships with the store names, for every dataset, its licence, its archive identifier and the reference to cite. A pretraining run therefore knows the terms of every sample it reads, and a user who needs a particular set of terms selects the buckets that meet them. Of the 77 datasets, 52 are CC0, 11 CC-BY 4.0, 12 CC-BY-SA 4.0 (including the nine Healthy Brain Network releases), one GNU General Public License v3.0 (GPL-3.0; Shin2017A), and one CC-BY-NC 4.0 (SparrKULee). Together, the 75 datasets under CC0, CC-BY, or CC-BY-SA hold 98.7\,\% of the hours (Fig.~\ref{fig:bubbles}b). Any researcher who uses the whole store is bound by each of these dataset terms (attribution for the 25 non-CC0 datasets, ShareAlike for 12, GPL-3.0 for one, and non-commercial use for SparrKULee), whereas leaving out the buckets of those last two datasets removes the GPL-3.0 and non-commercial terms. The repository is gated so that an account accepts these terms before downloading, and we claim no legal assurance beyond them; the terms are those of the source deposits, as the archives publish them.

\section{Does the Corpus Pretrain REVE-Base?}
\label{sec:validation}

To test whether the store supports pretraining, we retrain REVE-Base with its released trainer and recipe, then compare it with the released encoder on public tasks absent from the corpus.

\subsection{Retraining With REVE's Trainer}

To test whether \corpus{} pretrains REVE-Base without changing the architecture or objective, we train the model from scratch using REVE's released code and recipe. The objective is masked autoencoding of 10-second samples with 55\,\% block masking over channels and time. Batches are grouped by channel count ($4800/C$ samples per GPU) under StableAdamW with fp16 autocast and a trapezoid learning-rate schedule repeated each epoch. We modified the code only where it loads data, seeds, logs, and resumes (Appendix~\ref{app:patches}). In the released code, no seed is set and each GPU shuffles batches with its own unsynchronised generator, so an epoch on 16 GPUs covered only about 64\,\% of the batches (some of them twice). With our patch, both the weights and the batch order are seeded, and every epoch is one full pass over the corpus.

\begin{figure}[t]
\centering
\generated{training_loss.pdf}{\includegraphics[width=\columnwidth]{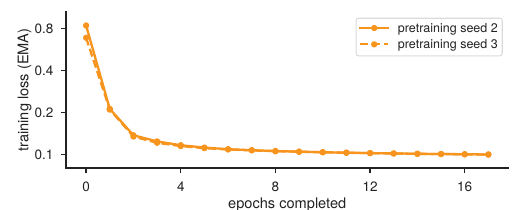}}
\caption{\textbf{For both seeds, the pretraining loss falls 3--4$\times$ within the first epoch and halves again over the next 16.} Reconstruction loss of REVE-Base on the corpus for seeds 2 (solid) and 3 (dashed), log scale. Epoch 0 is the untrained model's loss on its first batch; each later point is the mean over the epoch of the loss's exponential moving average.}
\label{fig:training}
\end{figure}

We train REVE-Base (22 layers, width 512, 69\,M parameters) twice from scratch for the 17 epochs and peak learning rate ($2.4\times10^{-4}$) of REVE's paper, using two distinct random seeds for weight initialisation and batch order (seeds 2 and 3). The first run read a store that also held I-CARE and excluded I-CARE through the loader argument of Section~\ref{sec:store} (23.5 hours on 16 A40 GPUs), whereas the second read the released store directly (29.7 hours on 16 V100 GPUs). Both runs read the same 632 buckets, and for both seeds the reconstruction loss falls sharply within the first epoch and then stabilises (Fig.~\ref{fig:training}).

\subsection{Scoring on Public Tasks}

\begin{figure*}[t]
\centering
\generated{results_grid.pdf}{\includegraphics[width=\textwidth]{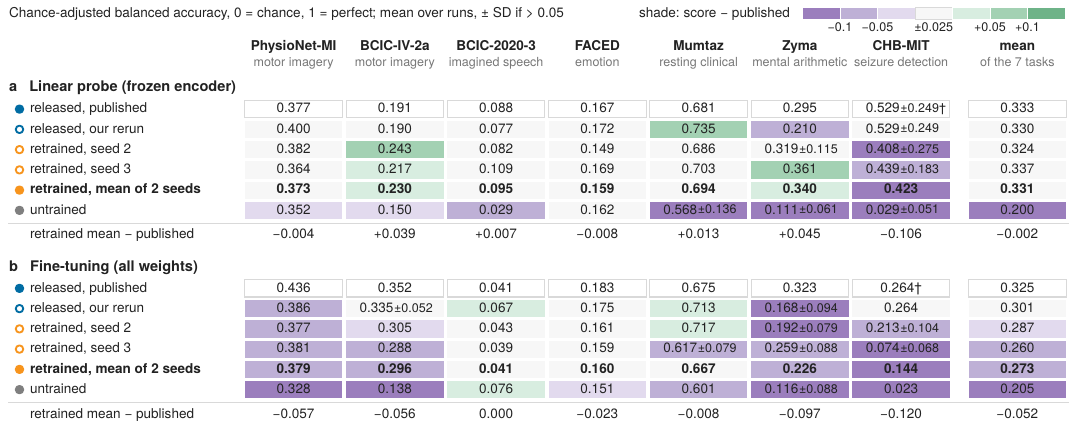}}
\caption{\textbf{Frozen, the retrained encoder matches the released one within its two-seed spread; fine-tuned, it falls 0.052 below the published scores and 0.028 below our rerun of the released weights.} Chance-adjusted balanced accuracy (0 = chance) on seven public OpenEEGBench tasks. Each task cell is the mean over three downstream runs (ten for our rerun on Zyma), $\pm$ their SD where it exceeds 0.05; the bold row is the mean of the two pretraining seeds and the last column the mean of the seven tasks. Shading is the cell minus the published score of the released REVE-Base in the same column (outlined row): purple below, green above, white within 0.025. $^\dagger$The leaderboard does not score CHB-MIT, so its reference is our rerun of the released weights. Last row of each block: retrained mean minus published.}
\label{fig:replication}
\end{figure*}

To evaluate the frozen and fine-tuned encoders under fixed downstream conditions, we score each checkpoint on seven public OpenEEGBench tasks (Fig.~\ref{fig:replication}). OpenEEGBench fits a linear head on the frozen encoder (the \emph{probe}) and, separately, fine-tunes every weight on each task's fixed split with one schedule and three seeds per cell. We report chance-adjusted balanced accuracy, $(\mathrm{BA}-1/K)/(1-1/K)$ for $K$ classes (0 = chance). Eight of the benchmark's twelve tasks use open datasets (the other four are gated). Seven of these eight tasks test paradigms represented in the corpus: motor imagery (BCIC-IV-2a, PhysioNet-MI), imagined speech (BCIC-2020-3), emotion (FACED), resting clinical EEG (Mumtaz), seizure detection (CHB-MIT), and mental arithmetic (Zyma, represented in the corpus by 15 hours of serial subtraction~\cite{wang_ds004148} and by the arithmetic problems between study and recall in PEERS ltpFR2 sessions). Because the remaining open task (ISRUC) is sleep staging and the corpus holds no sleep recordings, we leave it out. For the six tasks scored on the public leaderboard, the reference is the published score of the released REVE-Base, whereas for CHB-MIT (which the leaderboard does not score) the reference is our own run of the released weights through the same pipeline. On the seven tasks the leaderboard scores for the released model (the six open tasks above plus ISRUC, a set that differs from Fig.~\ref{fig:replication}), our rerun of the released weights reproduces its probe mean (0.346 vs.\ 0.349) but not its fine-tuning mean (0.352 vs.\ 0.376), because the package ships one fine-tuning schedule and the leaderboard's schedule is not published.

\subsection{The Frozen Encoder Reaches the Released Model's Score Within Seed Spread}

To test whether the frozen representations learned from the public corpus match those of the released model, we first compare linear-probing scores across the seven tasks (Fig.~\ref{fig:replication}). Over the seven tasks, the retrained model reaches a mean of 0.331 (seeds 0.324 and 0.337), compared with 0.333 for the released model and 0.200 for the untrained encoder. The difference of $-0.002$ lies within the 0.013 spread between the two pretraining seeds. Nor does this mean hide large discrepancies across individual tasks: on six of the seven tasks the two models differ by less than 0.05, on four of them by no more than the seed spread, and on BCIC-IV-2a ($+0.039$) and Zyma ($+0.045$) the retrained model scores higher. The sole exception is CHB-MIT, where the retrained model reaches 0.423 against 0.529. On that task, the SD over the three downstream runs of either model is 0.18 to 0.28, larger than the difference between the two encoders.

\subsection{After Fine-Tuning, a Gap Remains on Four Tasks}

We next fine-tune all encoder weights on the same seven tasks (Fig.~\ref{fig:replication}). After fine-tuning, the retrained model reaches 0.273 (seeds 0.287 and 0.260), compared with 0.325 published and 0.301 for our rerun of the released weights (untrained: 0.205). Its difference of $-0.052$ from the published mean exceeds the 0.027 seed spread and arises from four tasks (PhysioNet-MI at $-0.057$, BCIC-IV-2a at $-0.056$, Zyma at $-0.097$, and CHB-MIT at $-0.120$), whereas on the remaining three tasks the models differ by less than 0.025. Part of this gap belongs to the scoring pipeline. Rerun with the package's fine-tuning schedule, the released weights lose 0.024 against their published mean, leaving a difference of $-0.028$ against a like-for-like reference. In both regimes, the two retrained seeds remain above the untrained encoder on every task except FACED under probing (0.149 for seed~2 vs.\ 0.162) and BCIC-2020-3 under fine-tuning (where no model exceeds 0.08). Pretraining on the 77 redistributable datasets replicates the frozen representations of the released REVE-Base encoder across the seven public tasks, whereas a gap remains under full-network fine-tuning.

\section{Discussion}
\label{sec:discussion}

We released the redistributable part of REVE's corpus so that groups can fetch it by name and train on it directly. Trained from scratch on 14,075 public hours, REVE-Base learns a frozen encoder within the seed spread of the released model (0.331 vs.\ 0.333), which was trained on 4.4 times as many hours (61,415 hours). First, we isolated and converted the 77 redistributable datasets of REVE's corpus (6.21 billion tokens, or 79\,\% of the paper's tokens). Second, we packaged them into 632 dataset-scoped Zarr buckets (3.6\,TB) that keep each dataset's licence and allow any subset to be selected by name. Finally, we retrained REVE-Base from scratch and benchmarked both frozen and fine-tuned encoders across seven public tasks on OpenEEGBench (Fig.~\ref{fig:replication}). Which datasets enter, which recipe converts them, and which model reads them thus become choices that any group can vary while checking every file against its checksum and every dataset against its licence.

\textbf{What the fine-tuning gap leaves open.} After fine-tuning, the retrained model falls 0.052 below the published scores (on the two motor-imagery tasks, Zyma, and CHB-MIT) and 0.028 below our like-for-like rerun. These results suggest that two factors may account for the remaining gap, and our data do not separate them. The larger clinical corpus may yield weights that adapt better once every layer moves, or the package's fine-tuning schedule may favour the released weights. We also leave open whether the dense research-cap tokens of the corpus are worth as much downstream as the clinical tokens behind the released model. Because selecting a subset of datasets requires only a single loader argument, each of these questions is now an experiment that the corpus makes directly testable.

\textbf{Scope.} The store uses one recipe (10-second windows of 2,000 samples at 200\,Hz band-passed at 0.5--99.5\,Hz with per-session statistics beside them), and only REVE-Base was trained from it here. Other models may require different filtering, referencing, normalisation, channel handling, or windowing, even at 200\,Hz. Shorter patches can be cut within a window (although direct compatibility with LaBraM, CBraMod, or BIOT has not been tested), whereas longer contexts require checks of session continuity and other frequency bands require reconversion from the raw recordings. Second, two pretraining seeds bound the run-to-run variation by a range rather than a distribution, and they do not establish statistical equivalence to the released model. Finally, REVE's token count is estimated from its paper's channel-range table rather than counted directly from its private training files.

\textbf{The next version.} The three components of this release (a catalogue that resolves names to files, a converter that applies a recipe with an audit report, and a store that a trainer reads directly) carry over to other datasets and recipes. Memory and cognition holds 69\,\% of the current hours, with only one seizure dataset and no sleep or intracranial recordings. The next version therefore adds the public, non-gated EEG datasets of the EEGDash catalogue that REVE's corpus does not name to span those paradigms. A recipe step in braindecode can store a chosen sampling rate, band, window, and normalisation in the same layout (for instance, a 0.1\,Hz high-pass with no other irreversible step to serve the input pipelines of REVE, LaBraM, CBraMod, BIOT, and EEGPT from one set of bytes). Streaming samples directly from the hosted store during training will also remove the 3.6\,TB local download.

A public, training-ready EEG corpus suffices to pretrain the frozen representations of a foundation model. Dataset-scoped licences and subset selection give future releases a base to extend.

\section*{Acknowledgments}
This work used Expanse at the San Diego Supercomputer Center and Delta and DeltaAI at the National Center for Supercomputing Applications through allocations CIS261538, CIS262020 and IBN140002 from the Advanced Cyberinfrastructure Coordination Ecosystem: Services \& Support (ACCESS) program, which is supported by U.S. National Science Foundation grants \#2138259, \#2138286, \#2138307, \#2137603 and \#2138296. We thank the authors of REVE for releasing their code, weights and release card, and the authors of OpenEEGBench for the benchmark and its leaderboard.

\bibliographystyle{IEEEtran}
\bibliography{references}

\setcounter{topnumber}{2}\setcounter{bottomnumber}{2}\setcounter{totalnumber}{4}\renewcommand{\topfraction}{0.9}\renewcommand{\bottomfraction}{0.9}\renewcommand{\textfraction}{0.05}
\appendices

\section{REVE's paper, its release and the corpus}
\label{app:compare}

Table~\ref{tab:compare} compares REVE's paper, its released copy, and the corpus. Two thirds of the release's 34,400 hours come from I-CARE's 19-channel monitoring, so the corpus holds 41\,\% of the release's hours but 94\,\% of its tokens. We estimate the paper's token count from its channel-range table as hours $\times$ 3,600 $\times$ 1.1 $\times$ mean channels. REVE's own compute estimate assumes 68 mean channels, above that table's maximum of 46, whereas the other counts come directly from the converted samples.

\begin{table}[h!]
\centering
\caption{REVE's pretraining corpus as reported in its paper, REVE's released copy and the corpus.}
\label{tab:compare}
\footnotesize
\setlength{\tabcolsep}{4pt}
\begin{tabular}{@{}lrrrr@{}}
\toprule
corpus & datasets & hours & mean ch. & tokens \\
\midrule
REVE paper~\cite{elouahidi2025reve} & 92 & 61,415 & $\sim$32 & $\sim$7.9\,B \\
REVE's release~\cite{reve_dataset_card} & 70 & 34,400 & 49 & 6.59\,B \\
\corpus{} & 77 & 14,075 & 112 & 6.21\,B \\
\bottomrule
\end{tabular}
\end{table}

\section{Changes to the training code}
\label{app:patches}

We made six changes to REVE's released training code, none of which touches the model, the objective or the optimiser. In data loading: (i)~a channel set that occurs in a single session is kept by the segment builder; (ii)~a data path ending in \texttt{.zarr} is read from the store, with one \texttt{pread} per sample, a batch's samples read in parallel threads, one open-file budget per process, and reads completed to the chunk's length; (iii)~a list of datasets to leave out is applied to the bucket index before batching; (iv)~the batch order is seeded and shared by every GPU, and the weights are seeded, so that an epoch is one pass over the corpus and a run can be repeated (the released code shuffles per GPU with an unsynchronised generator, so an epoch on 16 GPUs covered about 64\,\% of the batches). In the training loop: (v)~the Accelerate wrapper's type check is removed for multi-GPU runs, the loss is written per epoch, encoder snapshots are saved every $N$ epochs and logging occurs every 100 iterations; (vi)~a resumed run continues at the epoch after the last saved state instead of restarting at epoch 0. With these changes, REVE's loader reads the store at close to the speed of the memory-mapped layout it replaces. On 65 of the datasets, a training step on four GPUs then takes 0.92 seconds, against 0.80 with the memory-mapped layout and 3.61 when samples are read one by one through the Zarr library.

\generated{datasets_table.tex}{
\begin{table*}[t]
\centering
\scriptsize
\caption{The 77 datasets of the corpus, by paradigm and hours. The identifier names the archive (ds, OpenNeuro; nm, NEMAR). Paradigms: BCI, BCI (MI \& P300); perception, perception \& emotion; resting, resting \& consciousness; learning, learning \& reward; development, development (rest \& tasks); auditory, auditory \& language; memory, memory \& cognition; clinical, clinical routine EEG. Hours after conversion to 200\,Hz. Short names of the MOABB datasets are MOABB's.}
\label{tab:datasets}
\setlength{\tabcolsep}{3pt}
\begin{tabular}{@{}lllrrl@{\hskip 14pt}lllrrl@{}}
\toprule
id & short name & paradigm & subj. & hours & licence & id & short name & paradigm & subj. & hours & licence \\
\midrule
nm000338 & Lee2019-MI & BCI & 54 & 183 & CC0 & ds002680 & Go/no-go & perception & 14 & 21 & CC0 \\
nm000250 & Dreyer2023 & BCI & 87 & 127 & CC BY & ds004324 & Toon faces & perception & 26 & 19 & CC0 \\
nm000323 & Lee2019ERP & BCI & 54 & 116 & CC0 & ds004043 & Feature attention & perception & 20 & 18 & CC0 \\
nm000267 & Shin2017A & BCI & 29 & 29 & GPL-3.0 & ds005594 & Alphabetic decision & perception & 16 & 12 & CC0 \\
nm000172 & Schirrmeister2017 & BCI & 14 & 29 & CC BY & ds004817 & RSVP attention 2 & perception & 20 & 12 & CC0 \\
nm000173 & Ofner2017 & BCI & 15 & 27 & CC BY & ds004357 & Features-EEG & perception & 16 & 11 & CC0 \\
nm000217 & BI2015b & BCI & 44 & 26 & CC BY & ds004816 & RSVP attention 1 & perception & 20 & 11 & CC0 \\
nm000245 & Cho2017 & BCI & 52 & 20 & CC0 & ds004395 & PEERS & memory & 364 & 9,111 & CC0 \\
nm000266 & Sosulski2019 & BCI & 13 & 20 & CC BY-SA & ds004706 & Spatial memory & memory & 34 & 471 & CC0 \\
nm000146 & Weibo2014 & BCI & 10 & 13 & CC0 & ds005121 & Siefert2024 & memory & 34 & 41 & CC0 \\
nm000244 & BI2014a & BCI & 64 & 12 & CC BY & ds005189 & Recollection & memory & 30 & 20 & CC0 \\
nm000216 & BI2015a & BCI & 43 & 12 & CC BY & ds005095 & Sternberg & memory & 48 & 17 & CC0 \\
nm000158 & Liu2024 (stroke) & BCI & 59 & 9 & CC BY & ds004951 & Braille letters & learning & 11 & 26 & CC0 \\
nm000215 & BI2014b & BCI & 38 & 2 & CC BY & ds003846 & Prediction error & learning & 19 & 23 & CC0 \\
nm000138 & AlexMI & BCI & 8 & 1 & CC BY-SA & ds004477 & Pandemic scenario & learning & 9 & 14 & CC0 \\
nm000273 & Lee2019SSVEP & SSVEP & 54 & 41 & CC0 & ds004152 & Drum trainer & learning & 21 & 11 & CC0 \\
nm000123 & Kalunga2016 & SSVEP & 12 & 3 & CC BY & ds004561 & Agency illusion & learning & 23 & 11 & CC0 \\
ds004148 & Test-retest rest & resting & 60 & 75 & CC0 & ds004147 & Task value & learning & 12 & 10 & CC0 \\
ds003969 & Meditation vs think & resting & 98 & 67 & CC0 & ds004284 & Neuroforecasting & learning & 18 & 9 & CC0 \\
ds004572 & Hypnosis & resting & 52 & 53 & CC0 & ds004262 & Feedback & learning & 21 & 8 & CC0 \\
ds003690 & Rest \& auditory & resting & 75 & 48 & CC0 & ds004264 & Steer the ship & learning & 21 & 7 & CC0 \\
ds001787 & Meditation & resting & 24 & 28 & CC0 & nm000238 & SparrKULee & auditory & 80 & 156 & CC BY-NC \\
ds004902 & Sleep deprivation & resting & 71 & 18 & CC0 & ds004279 & Spanish sentences & auditory & 56 & 54 & CC0 \\
ds005620 & Anaesthesia & resting & 21 & 18 & CC0 & ds004356 & Music vs speech & auditory & 22 & 46 & CC0 \\
ds003775 & SRM rest & resting & 110 & 10 & CC0 & ds005262 & Arabic inner speech & auditory & 12 & 25 & CC0 \\
ds002778 & Parkinson rest & resting & 31 & 3 & CC0 & ds004408 & Natural speech & auditory & 19 & 21 & CC0 \\
nm000232 & THINGS-EEG2 & perception & 10 & 87 & CC BY & ds004840 & Music therapy & auditory & 9 & 11 & CC0 \\
ds005697 & Perceive/imagine & perception & 45 & 69 & CC0 & ds005511 & HBN R7 & development & 381 & 280 & CC BY-SA \\
ds005089 & Selective attention & perception & 33 & 63 & CC0 & ds005508 & HBN R4 & development & 324 & 262 & CC BY-SA \\
ds005273 & Seen vs unseen & perception & 32 & 53 & CC0 & ds005509 & HBN R5 & development & 330 & 255 & CC BY-SA \\
ds003004 & Imagined emotion & perception & 34 & 48 & CC0 & ds005514 & HBN R9 & development & 295 & 211 & CC BY-SA \\
ds003825 & THINGS-EEG1 & perception & 50 & 46 & CC0 & ds005512 & HBN R8 & development & 257 & 179 & CC BY-SA \\
ds004582 & Fake faces & perception & 73 & 34 & CC0 & ds005507 & HBN R3 & development & 184 & 159 & CC BY-SA \\
ds005586 & Numerosity & perception & 23 & 34 & CC0 & ds005506 & HBN R2 & development & 150 & 128 & CC BY-SA \\
ds004603 & Context trajectory & perception & 37 & 31 & CC0 & ds005505 & HBN R1 & development & 136 & 118 & CC BY-SA \\
ds003885 & Object motion 1 & perception & 24 & 27 & CC0 & ds005510 & HBN R6 & development & 135 & 103 & CC BY-SA \\
ds003887 & Object motion 2 & perception & 24 & 27 & CC0 & nm000181 & NMT & clinical & 2,413 & 489 & CC BY-SA \\
ds002721 & Affective music & perception & 31 & 26 & CC0 & nm000385 & Siena & seizure & 14 & 141 & CC BY \\
ds004315 & Mood induction & perception & 50 & 21 & CC0 &  & & & & &  \\
\midrule
\multicolumn{12}{@{}l}{\textbf{Total: 77 datasets, 7,272 subjects, 14,075 h}} \\
\bottomrule
\end{tabular}
\end{table*}
}

\newpage
\section{The 77 datasets}
\label{app:datasets}

Table~\ref{tab:datasets} lists every dataset with its paradigm, subjects, hours, and licence. PEERS is stored with 125 to 256 channels. The other 46 OpenNeuro datasets are research studies of perception, attention, memory, resting state, emotion, and language, with 8 to 229 channels. The other 20 NEMAR datasets hold motor-imagery, P300, SSVEP, auditory, perception, and seizure recordings, with 8 to 128 channels. NMT contributes 2,413 of the 7,272 subjects and the Healthy Brain Network 2,192; these are distinct participants, since the participant tables of its nine releases share no identifier.

\section{Datasets of REVE's corpus left out}
\label{app:excluded}

Table~\ref{tab:excluded} lists the 13 datasets named in REVE's paper that are not in the corpus, and the three entries of REVE's release card that it leaves out. The card's Healthy Brain Network entry resolves to nine releases rather than ten: EEGDash holds one release twice (as a 100\,Hz challenge copy and as the 500\,Hz accession), and the corpus takes the 500\,Hz one.

\begin{table}[t]
\centering
\caption{Datasets named by REVE's paper or release card that the corpus leaves out.}
\label{tab:excluded}
\scriptsize
\setlength{\tabcolsep}{3pt}
\begin{tabular}{@{}>{\raggedright\arraybackslash}p{4.3cm}>{\raggedright\arraybackslash}p{4.1cm}@{}}
\toprule
dataset & reason \\
\midrule
TUH EEG~\cite{obeid2016tuh}; Harmful Brain Activity; TDBRAIN; I-CARE~\cite{amorim2023icare} & registration or a data use agreement~\cite{bdsp_icare}; about 50,000 of REVE's 61,415 hours \\
BNCI2014004/008/009, BNCI2015001/003/004, Nakanishi2015, EPFL P300 (MOABB) & no-derivatives terms (six datasets) or no explicit redistribution licence (two); excluded under this release's policy \\
Zhou2016 (MOABB)~\cite{zhou2016mi} & the card's deposit (\texttt{nm000226}) is a 100\,Hz copy, below the 200\,Hz floor; the 250\,Hz original (\texttt{nm000115}) can enter a later version \\
\midrule
PhysioNet MI~\cite{schalk2004bci2000}, \texttt{ds004362}; \texttt{ds005597}; ``other PhysioBank items'' & on the card only: a test task; fMRI without EEG; no dataset named \\
\bottomrule
\end{tabular}
\end{table}

\end{document}